\documentclass[12pt]{article}

\long\def\nop#1{}

\def\comment{\edef\cps{\the\parskip} \parskip=0.5cm \begingroup \tt}
\def\endcomment{\endgroup \vskip 0.5cm \parskip=\cps}

\expandafter\ifx\csname shortcite\endcsname\relax

\fi

\newbox\current

\long\def\plframebox#1{
\setbox\current\vbox{#1}		

\vbox to \ht\current {\hrule\vss
\hbox to \wd\current {%
\vrule \hss\box\current\hss \vrule}
\vss\hrule }
}

\long\def\eatpar#1{%
\ifx#1\par                      
\let\nextmove=\eatpar           
\else
\let\nextmove=#1
\fi
\noexpand\nextmove
}

\def\modifymargins#1#2{
\newdimen\addtoh
\newdimen\addtow
\addtoh=#1
\addtow=#2

\advance\topmargin by -\addtoh
\multiply\addtoh by 2
\advance\textheight by \addtoh

\advance\oddsidemargin by -\addtow
\advance\evensidemargin by -\addtow
\multiply\addtow by 2
\advance\textwidth by \addtow
}

\begingroup
\catcode`\~=11
\gdef\centertilde#1{\lower #1pt\hbox{~}}
\endgroup

\newcount\currenttime
\newcount\hour
\newcount\minute

\def\printtime{%
\currenttime=\time
\hour=\currenttime
\divide\hour by 60
\minute=-\hour
\multiply\minute by 60
\advance\minute by \currenttime
\the\hour:\ifnum\minute<10 0\fi\the\minute
}

\begingroup
\makeatletter
\global\let\@@date=\@date
\gdef\@date{\@@date\ --- \printtime}
\endgroup

\def\oggi{\number\day\space 
\ifcase\month\or
Gennaio\or Febbraio\or Marzo\or Aprile\or Maggio\or Giugno\or
Luglio\or Agosto\or Settembre\or Ottobre\or Novembre\or Dicembre\fi
\space \number\year}

\newcounter{rmexample}

\def\proof{\noindent {\sl Proof.\ \ }}

\def\qed{\hfill{\boxit{}}
  \ifdim\lastskip<\medskipamount \removelastskip\penalty55\medskip\fi}
\def\qedn#1{\hfill{\boxit{}$_#1$}
  \ifdim\lastskip<\medskipamount \removelastskip\penalty55\medskip\fi}
\long\def\boxit#1{\vbox{\hrule\hbox{\vrule\kern3pt
                  \vbox{\kern3pt#1\kern3pt}\kern3pt\vrule}\hrule}}

\def\I{{\cal I}}  \def\K{{\cal K}} \def\L{{\cal L}}
\def\M{{\cal M}}   
 \def\R{{\cal R}}  \def\T{{\cal T}}

\def\ie{i.e.}
\def\eg{e.g.}

\def\M{{\cal M}}

\def\l{\langle}
\def\r{\rangle}

\def\mod{M\!od}

\def\var{V\!ar}

\def\true{{\sf true}}

\def\np{{\rm NP}}

\def\P#1{\mbox{$\Pi^p_{#1}$}}

\def\Dlog#1{\mbox{$\Delta^p_{#1}[\log n]$}}

\def\profont{\sf}

\def\x3c{{\profont x3c}}

\def\possnewtheorem#1#2{
\expandafter\ifx\csname #1\endcsname\relax
\newtheorem{#1}{#2}
\fi
}

\def\possnewtheoremthree#1[#2]#3{
\expandafter\ifx\csname #1\endcsname\relax
\newtheorem{#1}[#2]{#3}
\fi
}

\possnewtheorem{theorem}{Theorem}
\possnewtheorem{corollary}{Corollary}
\possnewtheorem{lemma}{Lemma}
\possnewtheoremthree{proposition}[theorem]{Proposition}
\possnewtheorem{definition}{Definition}
\possnewtheorem{question}{Question}
\possnewtheorem{example}{Example}
\possnewtheorem{nontheorem}{Counterexample}
\possnewtheorem{property}{Property}
\possnewtheorem{assumption}{Assumption}
\possnewtheorem{conjecture}{Conjecture}
\newenvironment{theorem*}[1]{{\noindent \bf Theorem~#1}\begin{it}}{\end{it}\

}

\let\texcirc=\circ
\def\circ{C\!I\!RC}
\def\ncirc{\mbox{{\sc ncirc}}}

\let\verbose=\iffalse
\let\endverbose=\fi

\long\def\comment#1\endcomment{}

\modifymargins{20pt}{20pt}
\long\def\comment#1\endcomment{}

\title{Merging Locally Correct Knowledge Bases:\\
A Preliminary Report}

\author{Paolo Liberatore\\
\normalsize Dipartimento di Informatica e Sistemistica \\
\normalsize Universit\`a di Roma ``La Sapienza'' \\
\normalsize Via Salaria 113, 00198, Roma, Italy \\
\normalsize Email: {\tt paolo@liberatore.org} \\
}

\date{}

\begin{document}

\maketitle

\begin{abstract}

Belief integration methods are often aimed at
deriving a single and consistent knowledge base that retains
as much as possible of the knowledge bases to integrate.
The rationale behind this approach is the {\em minimal change
principle}: the result of the integration process should differ
as less as possible from the knowledge bases to integrate. We show that
this principle can be reformulated in terms of a more general
model of belief revision, based on the assumption that
inconsistency is due to the {\em mistakes} the knowledge bases
contain. Current belief revision strategies are based on a
specific kind of mistakes, which however does not include all
possible ones. Some alternative possibilities are discussed.

\end{abstract}


\comment
Note that A and B are not IC, as they are used for
a different purpose (not to enforce something that must
be true in general, but to check whether the result of
merging looks reasonable: the difference is that the IC
are often fixed, while A and B may change at each merging).
However, they also may include IC.
\endcomment

\section{Introduction}
\label{intro}

Most of the existing belief revision semantics are based---in
some way---on a principle that has been formulated at the
very beginning of the investigation on this topic: the
{\em minimal change principle} \cite{alch-gard-maki-85,gard-88}.
According to this principle, the result of integrating two
or more knowledge bases should be as similar as possible to
them. Semantics proposed for merging agree on this principle,
and only differ in the way it is applied, \ie, in how to
combine the several possibilities arising, in how to measure
the difference between knowledge bases, in how the knowledge
bases are represented, in what is the relative reliability
of sources, etc. Nevertheless, very few arguments against
the basic principle exist \cite{rott-00}.

This paper does not contain arguments against the minimal
change principle, but only as it being a first principle. 
Taking a different perspective, we show that
it is indeed a particular consequence of a more general
assumption. Namely, we present a model of how the knowledge
bases to integrate are obtained that justifies the minimal
change principle, as it is currently applied, only in
particular cases. This model explains inconsistencies
between knowledge bases by assuming that mistakes have
been done in the process of knowledge acquisition.

This model is not completely new, as existing merging
semantics actually rely on its particularization to
the case in which mistakes are changes of value
of literals. For example, Dalal's revision semantics
\cite{dala-88} can be reformulated as the result of
assuming that one knowledge base
is free of mistakes, and the other one results from introducing
mistakes in the value of literals in an otherwise correct
knowledge base. In formulae, while revising $K$ with $P$,
we assume that the process of acquiring $P$ is error-free, while
$K$ contains some mistakes, each changing
the value of a single literal in a model. Therefore,
Dalal's revision can be reformulated as the correction
of a minimal number of mistakes. Other belief revision
semantics are based on the same principle, but have different
rules for combining the different possibilities that arise
\cite{sato-88,forb-89,borg-85}. Iterated belief revision
semantics \cite{will-94,libe-97-c}, updates
\cite{libe-97,herz-rifi-99,herz-etal-01},
and merging/arbitration operators
\cite{libe-scha-98-b,koni-pere-98,libe-scha-00-c,meye-etal-02},
are based on similar principles.

The model proposed in this paper, however, does not only formalize
existing semantics; being more general, it is applicable
to other scenarios, leading to different revision techniques.
While cases like the
example of the stock market experts~\cite{koni-pere-98}
are perfectly modeled in the ``mistake of value'' model,
other ones are not. Some examples, like the following one,
comes from everyday life.

\begin{example}

Yesterday, I met an old friend I have not been seeing in
years. While talking about the high school days, we
shared information about other friends we knew at that
time. In particular, he told me that George earned a
lot of money by creating a startup company he then sold,
and now he lives in the Nukunonu island. On the other hand,
I knew that George become incredibly rich with some
illegal business, and he is currently in jail (but I do not
know whether he still has some of the money.)

The union of our knowledge bases is inconsistent, as there are
no jails in the Nukunonu island. On the other hand, both of us are
completely certain of our current knowledge. We then had
to conclude that we were talking about two different Georges.
The correct conclusion of merging information should then be that
``George\_A is rich'', ``George\_A lives in the Nukunonu island'',
and that ``George\_B is in jail''.

Merging based on the minimal change principle, combined with
the ``mistake of value'' assumption as it is usually done,
would have led to a completely different result. Namely,
since we assumed that we are talking about the same George,
and since both of us have the same confidence on our knowledge,
we could only conclude that either ``George is in jail'' or that
``George lives in the Nukunonu island'', but not both (since
no jail is in the Nukunonu island.) This is
already a problem, as this information is not complete about
George's current location, while in fact we both {\em know}
exactly where the Georges are. Still worst, since I do not
know whether George is still rich while my friend is sure
he is, I will incorrectly conclude that the George I am
talking about is still rich, a fact that is not backed up
by any evidence.

\end{example}

This scenario is about a common life incident, but
similar problems are common in computer science:
putting together two \LaTeX\  source files creates the problem of
the same name for two different macros; similar problems
arise in compiling C code fragments, etc. In the rest of
the paper, we make the simplifying assumption that each
knowledge base is the knowledge of a different agent involved
in the process of merging.

One of the characteristics of the example above is
the ``local'' correctness of the involved knowledge
base: both me and my  friend had correct information
about the George we were thinking about. The fact that
each agent regards its knowledge base as correct, and
then has to correct it during the merging process, is
true in current belief semantics as well.
However, the ``mistake of value'' model
implies that the conclusions drawn by each single knowledge
base were in fact incorrect. On the contrary, if the only
mistakes are like the same name for two different objects,
then the conclusions drawn from each knowledge base
separately (before the merging) are correct, \eg, the
conclusion that George cannot travel any more was correctly
entailed by my knowledge base, and this is a correct conclusion,
as I am referring to the George who is in jail.
The correction to the knowledge bases
is therefore only necessary to avoid inconsistency while
merging the knowledge bases.

While inconsistency is undoubtedly the most serious
problem that may arise during merging, it is not
the only one. There are mistakes that cannot be be discovered
just by checking for inconsistency the union of the knowledge
bases. Indeed, a mistake does not necessarily create an
inconsistency. On the contrary, some mistakes make the
union of the knowledge bases weaker than it should be. An
example of this case is when two knowledge bases give different
names to the same object, which forbids drawing conclusions
based on two facts contained in the two knowledge bases.

\begin{example}

Still talking with my high school friend, I mentioned
Teddy, who entered the Law school; I though that if he
ever had graduated, he would have ended up in jail. The
friend I was talking with, however, does not remember this
Teddy, and the only guy he knows entered Law was Bobby,
who actually graduated. In fact, Teddy was a nickname
for Bobby, but we did not remember this fact.

No inconsistency arises in this case. However,
the conclusion that Bobby is (likely) in jail could not
be drawn by simply putting together the knowledge we had.
Contrary to common merging scenarios, the conjunction of
the knowledge bases is weaker than it should be. Such
problems are clearly difficult to diagnose, as they do not
create an inconsistency. The only way to find them out is
from the fact that the resulting knowledge base is weaker
than it should be. For example, knowing that only one
person from our class entered the Law school would have
allow us to find out that Bobby and Teddy must be the
same person.

\end{example}

In order to produce a knowledge base in which as many mistakes as
possible are corrected, we use two formulae that act as integrity
constraints. Formally, we are given a multiset of knowledge bases
$\K$ and two formulae $A$ and $B$; the result of the integration process
is a formula $K=\I^A_B \K$ such that $K \models A$ and
$K \wedge B \not\models \bot$. This way, we constraint the resulting
knowledge base to have (at least) a specific set of consequences $A$,
and not to have some undesired other consequences $B$. The formula
$A$ formalizes the usual integrity constraints (facts that should
remain true after integration), while $B$ extends the usual
consistency requirement: $B=\top$ only enforces the result of
integration to be consistent. Since
$K = \I^A_B \K$ is a formula whose set of models is contained
in $\mod(A)$, and is not contained in $\mod(\neg B)$, we call
$A$ and $B$ the upper and lower bound of the merging operator,
respectively.

If the union of the knowledge bases of $\K$ implies $A$ and
is consistent with $B$, we assume that there is no problem,
\ie, the knowledge bases do not contain any mistake.
This assumption may be wrong anyway, but we have no
way to realize it. The interesting case is when either constraint
is not satisfied. In this case, we assume that some mistakes
have been made while acquiring the knowledge bases. Some
possible mistakes are listed below. The three last mistakes
of the list are the only ones leading to a locally incorrect
knowledge base.

\begin{description}

\item[homonymy:] two agents use the same variable while
they should use two different ones;

\item[synonimies:] two agents use different variables while they
should use the same one;

\item[subject misunderstanding:] a formula is stated using
one variable, while it should use a different one;

\item[extension:] a formula $F$ is extended to another variable
or set of variables: formally, the agent assumes $F[X/Y]$
in addition of $F$;

\item[generalization:] a formula is assumed to hold in
general, while it holds only under some assumptions;

\item[particularization:] a formula is assumed to hold only in
a specific scenario, while it is more general;

\item[ambiguity:] a formula containing $a \vee b$ is assumed
to be $a$ alone (or $b$ alone, or both $a$ and $b$);

\item[exclusion:] a formula containing an inclusive or is taken
to refer to the exclusive or, or vice versa;

\item[value:] the formula is correct because it
contains a model with a wrong value.

\end{description}

Besides the mistake of value,
these mistakes can be grouped in three categories:
mistakes due to a wrong interpretation of variables (homonymies,
synonimies, and subject misunderstanding); mistakes due
to a wrong interpretation of context (generalization,
particularization, and extension); mistakes due to a
wrong of the logic (ambiguity and exclusion).

Some other mistakes are particular cases of the above ones.
For example, an agent may incorrectly assume that a previously
true fact continues to hold while it does not: this is a
subcase of incorrect generalization. Another similar mistake
is the incorrect simplification of a definition, like ``the
water boils at $100^\texcirc$C'' instead of ``the water boils
at $100^\texcirc$C at sea level''.

In the domain we consider, each agent introduces some mistakes
into a truly correct knowledge base. This is modeled by assuming
that each agent modified its original knowledge base in some
way. Clearly, this is only a theoretical model: if the agent ever had a
correct knowledge base, it had not modify it. However, this way
we can say that ``the agent modified the knowledge base'', that
simplifies the more correct sentence ``the agent incorrectly
considered the information $x$ to be~$y$''.

Merging is the process of first correcting mistakes in the
knowledge bases, and then conjoining them. Correcting mistakes,
in turns, is a two-phase process: first, we have to find out
which mistakes have been made, and then correcting them. We
initially assume that an ordering of likeliness of mistakes
is known, and then consider the problem of how to derive it
from the knowledge bases. In this second
case, however, we cannot expect the merging process to do much,
given the high number of possible mistakes: for example, the
multiset $\{ a, \neg a \}$ may be inconsistent because the
second $a$ should be $b$, or because $a$ is only true when
$b$ is true (that is, the first formula should be $b \rightarrow a$
instead of $a$ alone), or because the ambiguity $a \vee b$ of
the first formula has been interpreted as a choice, and the agent
has incorrectly assumed $a$, etc. The number of possibilities increases
with the number of variables and with the size and complexity of the
knowledge bases. The process of correcting the mistakes can
also be problematic: knowing that a formula has been obtained
by changing a name is not enough if we do not know the original
name.

We make some simplifications. The first one is to
neglect the mistakes of logic (ambiguity and exclusion).
The second one is to restrict our study to propositional
knowledge bases. While first-order logic (even without function
symbols) is uncommon in belief revision studies, this assumption
makes us disregard the very relevant case of epistemic bases
\cite{lang-marq-will-01}, which contain not only the agent's
belief, but also what it considers more or less plausible.

\section{A Model of the Sources}
\label{model}

\begin{figure}[ht]
\begin{center}
\setlength{\unitlength}{4144sp}%
\begingroup\makeatletter\ifx\SetFigFont\undefined%
\gdef\SetFigFont#1#2#3#4#5{%
  \reset@font\fontsize{#1}{#2pt}%
  \fontfamily{#3}\fontseries{#4}\fontshape{#5}%
  \selectfont}%
\fi\endgroup%
\begin{picture}(4345,1947)(708,-2176)
\thinlines
\put(991,-511){\framebox(720,270){}}
\put(2071,-511){\framebox(720,270){}}
\put(3061,-511){\framebox(720,270){}}
\put(4321,-511){\framebox(720,270){}}
\put(2611,-1681){\framebox(720,270){}}
\put(2431,-511){\line( 0,-1){540}}
\put(2431,-1051){\line( 1, 0){450}}
\put(2881,-1051){\vector( 0,-1){360}}
\put(3421,-511){\line( 0,-1){540}}
\put(3421,-1051){\line(-1, 0){360}}
\put(3061,-1051){\vector( 0,-1){360}}
\put(1351,-511){\line( 0,-1){720}}
\put(1351,-1231){\line( 1, 0){1350}}
\put(2701,-1231){\vector( 0,-1){180}}
\put(4681,-511){\line( 0,-1){720}}
\put(4681,-1231){\line(-1, 0){1440}}
\put(3241,-1231){\vector( 0,-1){180}}
\put(2971,-1681){\vector( 0,-1){270}}
\put(1351,-421){\makebox(0,0)[b]{\smash{\SetFigFont{12}{14.4}{\familydefault}{\mddefault}{\updefault}source 1}}}
\put(2431,-421){\makebox(0,0)[b]{\smash{\SetFigFont{12}{14.4}{\familydefault}{\mddefault}{\updefault}source 2}}}
\put(3421,-421){\makebox(0,0)[b]{\smash{\SetFigFont{12}{14.4}{\familydefault}{\mddefault}{\updefault}source 3}}}
\put(4681,-421){\makebox(0,0)[b]{\smash{\SetFigFont{12}{14.4}{\familydefault}{\mddefault}{\updefault}source n}}}
\put(2971,-1591){\makebox(0,0)[b]{\smash{\SetFigFont{12}{14.4}{\familydefault}{\mddefault}{\updefault}merger}}}
\put(1216,-961){\makebox(0,0)[rb]{\smash{\SetFigFont{12}{14.4}{\familydefault}{\mddefault}{\updefault}$K_1$}}}
\put(2341,-961){\makebox(0,0)[rb]{\smash{\SetFigFont{12}{14.4}{\familydefault}{\mddefault}{\updefault}$K_2$}}}
\put(3511,-961){\makebox(0,0)[lb]{\smash{\SetFigFont{12}{14.4}{\familydefault}{\mddefault}{\updefault}$K_3$}}}
\put(4771,-961){\makebox(0,0)[lb]{\smash{\SetFigFont{12}{14.4}{\familydefault}{\mddefault}{\updefault}$K_n$}}}
\put(2971,-2176){\makebox(0,0)[b]{\smash{\SetFigFont{12}{14.4}{\familydefault}{\mddefault}{\updefault}$K$}}}
\end{picture}
\caption{The basic model of belief merge.}
\label{model-basic}
\end{center}
\end{figure}
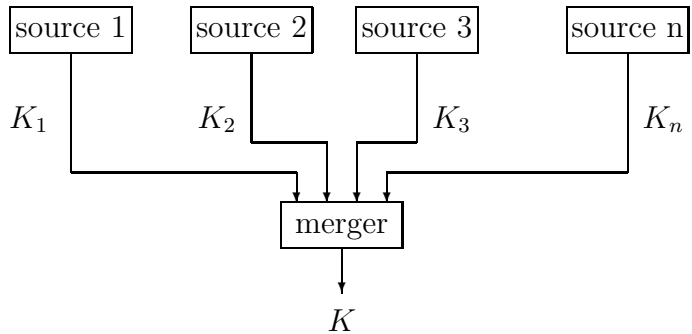

In this section, we give a formal definition of our
framework. The general belief merging process can be visualized
as in Figure~\ref{model-basic}: there are a number of agents
(sources) each sending a knowledge base to a centralized
``knowledge merger''. We do not consider the more sophisticated
models that are sometimes used (\eg, an agent supplies more
than one knowledge base.)

\begin{figure}[ht]
\begin{center}
\setlength{\unitlength}{4144sp}%
\begingroup\makeatletter\ifx\SetFigFont\undefined%
\gdef\SetFigFont#1#2#3#4#5{%
  \reset@font\fontsize{#1}{#2pt}%
  \fontfamily{#3}\fontseries{#4}\fontshape{#5}%
  \selectfont}%
\fi\endgroup%
\begin{picture}(4435,2880)(708,-2176)
\thinlines
\put(991,-511){\framebox(720,270){}}
\put(2071,-511){\framebox(720,270){}}
\put(4321,-511){\framebox(720,270){}}
\put(2611,-1681){\framebox(720,270){}}
\put(2431,-511){\line( 0,-1){540}}
\put(2431,-1051){\line( 1, 0){450}}
\put(2881,-1051){\vector( 0,-1){360}}
\put(3511,-511){\line( 0,-1){540}}
\put(3511,-1051){\line(-1, 0){450}}
\put(3061,-1051){\vector( 0,-1){360}}
\put(1351,-511){\line( 0,-1){720}}
\put(1351,-1231){\line( 1, 0){1350}}
\put(2701,-1231){\vector( 0,-1){180}}
\put(4681,-511){\line( 0,-1){720}}
\put(4681,-1231){\line(-1, 0){1440}}
\put(3241,-1231){\vector( 0,-1){180}}
\put(2971,-1681){\vector( 0,-1){270}}
\put(1351,299){\vector( 0,-1){540}}
\put(2431,299){\vector( 0,-1){540}}
\put(4681,299){\vector( 0,-1){540}}
\put(901,-691){\dashbox{57}(900,1170){}}
\put(1981,-691){\dashbox{57}(900,1170){}}
\put(3151,-511){\framebox(720,270){}}
\put(3511,299){\vector( 0,-1){540}}
\put(3061,-691){\dashbox{57}(900,1170){}}
\put(4231,-691){\dashbox{57}(900,1170){}}
\put(2971,-1591){\makebox(0,0)[b]{\smash{\SetFigFont{12}{14.4}{\familydefault}{\mddefault}{\updefault}merger}}}
\put(1216,-961){\makebox(0,0)[rb]{\smash{\SetFigFont{12}{14.4}{\familydefault}{\mddefault}{\updefault}$K_1$}}}
\put(2341,-961){\makebox(0,0)[rb]{\smash{\SetFigFont{12}{14.4}{\familydefault}{\mddefault}{\updefault}$K_2$}}}
\put(4771,-961){\makebox(0,0)[lb]{\smash{\SetFigFont{12}{14.4}{\familydefault}{\mddefault}{\updefault}$K_n$}}}
\put(2971,-2176){\makebox(0,0)[b]{\smash{\SetFigFont{12}{14.4}{\familydefault}{\mddefault}{\updefault}$K$}}}
\put(1216,119){\makebox(0,0)[rb]{\smash{\SetFigFont{12}{14.4}{\familydefault}{\mddefault}{\updefault}$S_1$}}}
\put(2341,119){\makebox(0,0)[rb]{\smash{\SetFigFont{12}{14.4}{\familydefault}{\mddefault}{\updefault}$S_2$}}}
\put(4771,119){\makebox(0,0)[lb]{\smash{\SetFigFont{12}{14.4}{\familydefault}{\mddefault}{\updefault}$S_n$}}}
\put(1351,-421){\makebox(0,0)[b]{\smash{\SetFigFont{12}{14.4}{\familydefault}{\mddefault}{\updefault}$\tau_1$}}}
\put(2431,-421){\makebox(0,0)[b]{\smash{\SetFigFont{12}{14.4}{\familydefault}{\mddefault}{\updefault}$\tau_2$}}}
\put(4681,-421){\makebox(0,0)[b]{\smash{\SetFigFont{12}{14.4}{\familydefault}{\mddefault}{\updefault}$\tau_n$}}}
\put(3511,-421){\makebox(0,0)[b]{\smash{\SetFigFont{12}{14.4}{\familydefault}{\mddefault}{\updefault}$\tau_3$}}}
\put(3556,119){\makebox(0,0)[lb]{\smash{\SetFigFont{12}{14.4}{\familydefault}{\mddefault}{\updefault}$S_3$}}}
\put(3601,-961){\makebox(0,0)[lb]{\smash{\SetFigFont{12}{14.4}{\familydefault}{\mddefault}{\updefault}$K_3$}}}
\put(1351,569){\makebox(0,0)[b]{\smash{\SetFigFont{12}{14.4}{\familydefault}{\mddefault}{\updefault}source 1}}}
\put(2431,569){\makebox(0,0)[b]{\smash{\SetFigFont{12}{14.4}{\familydefault}{\mddefault}{\updefault}source 2}}}
\put(3511,569){\makebox(0,0)[b]{\smash{\SetFigFont{12}{14.4}{\familydefault}{\mddefault}{\updefault}source 3}}}
\put(4681,569){\makebox(0,0)[b]{\smash{\SetFigFont{12}{14.4}{\familydefault}{\mddefault}{\updefault}source n}}}
\end{picture}
\caption{The model of belief merge, with mistakes.}
\label{model-understand}
\end{center}
\end{figure}
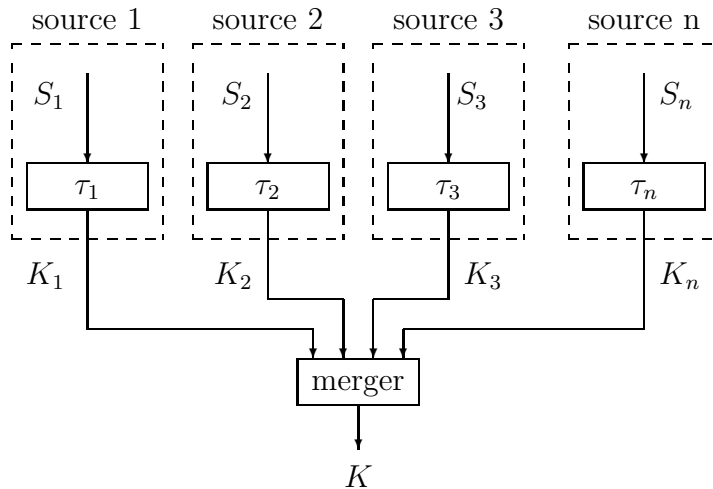

We improve over this simple schema by providing a model
of how the sources get the knowledge bases $K_i$'s they
pass to the merger: each $K_i$ is obtained
by applying one or more transformations to a knowledge base
$S_i$, which is assumed to be correct.
\verbose
In other words, we
start with a set of knowledge bases $\{S_i\}$; each of them
is modified by applying some transformation to it, thus
obtaining $K_i$. The multiset $\{K_i\}$ is what we have to merge.
\endverbose
Figure~\ref{model-understand} is a graphical representation
of this model.

The ``mistake of value'' revision semantics fit in this model: each
$K_i$ is obtained from $S_i$ by applying the transformation
that changes the value of a variable in a model. Namely, let
$\tau^v_{M,x}$ be the transformation that takes a formula, and gives
another formula in which the model $M$ is replaced by the model
with the opposite value of $x$. Each $K_i$ is obtained from
$S_i$ by applying a suitable number of such transformations.
Specific revision/arbitration/merging operators can be then
formalized by assuming a form of minimality of the mistakes,
and then combining in some way the possible results of this
assumption.

For example, Dalal's revision assumes that a.\  one of the
knowledge base is correct (no transformation has
been applied to it); b.\  the other knowledge base results from
the application of a number of transformations $\tau^v_{M,x}$
to a correct one; and c.\  a minimal number of transformations
have been applied. If more than one knowledge base result from
inverting these transformations, they are disjoined. This semantics
fits into the proposed model: the $K_i$'s are obtained by
applying transformations to the $S_i$'s, and the process of
integration attempts to invert them.

Formalizing Dalal's revision in this way shows how integration can be
done in general: inverting the transformation applied to $S_i$,
and merging what results. Ideally, we should be able to obtain
the knowledge bases $S_i$, which are assumed correct. 
Unfortunately, inverting the transformations cannot be done
uniquely, as the merger only knows the $K_i$'s, but has no direct
knowledge of the transformations used or the original $S_i$'s.
For example, $K_i=a$ may be correct, or may be the result of
changing a variable name to $S_i=b$, or may be a wrong
generalization of $S_i=c \rightarrow a$, and so on.

\verbose
Let us now be more precise about the transformations. These are
changes to the original knowledge bases $S_i$'s due to mistakes.
An informal list of the mistakes we consider are listed in
Section~\ref{intro}. Formally, we can state that each $\tau_i$
is the composition of one or more transformation in the list
below.
\endverbose

The mistakes listed in Section~\ref{intro} can
be formalized by the following transformations.

\begin{description}

\item[variable substitution:] $\tau^h_{x,y}(F) = F[x/y]$;

\item[generalization:] $\tau^g_{x}(F) = F[x/\true]$;

\item[particularization:] $\tau^p_{x}(F) = x \rightarrow F$;

\end{description}

Variable substitution models all mistakes due to mistakes relative
to variable names: homonymies, renaming, and subject misunderstanding.
\verbose
Each of them should be treated separately, as their effects are
quite different (homonymies may lead to inconsistency, while
renaming may weaken knowledge bases). However, they can all be
formalized using the same transformation.

Generalization and particularization have instead quite a
different definition.
\endverbose
Wrong generalization is the mistake of neglecting some
assumptions of an (otherwise true) fact.
This can be formalized by taking the original (correct)
formula $F$, and replacing the assumption $x$ with $\true$.
Note that the resulting formula $\tau^g_{x}(F)$ does not
contain $x$ at all, but has exactly the models $F$ would
have if $x$ is true.
The simplest case of generalization is
when $x \rightarrow F$ is taken to be $F$: if $F$ does not
contain $x$, then $F = \tau^g_{x}(x \rightarrow F)$. However,
$\tau^g_{x}$ also models more complex cases of generalization.
Particularization is easy to formalize: some
assumptions are believed to be required for some fact to hold,
while they are not.
Generalization and particularization
can be, to some extent, been considered the opposite of each
other, since $\tau^g_{x}( \tau^p_{x}(F) ) \equiv F$. However, the
converse does not hold, as it may be
$\tau^p_{x}( \tau^g_{x}(F) ) \not\equiv F$; this is the case,
for example, if $F$ does not mention $x$ at all.
We neglect mistakes of logic, that is, ambiguity
and exclusion, as they are too hard to detect and invert.
Especially ambiguity is difficult to detect without a lot
of additional information: given a formula,
it may be that each of its subformulae was originally disjoined
with another formula (that may be an arbitrary formula of
the domain). Even restricting to literals, the number of
possibilities makes the problem quite difficult.

\section{The Merging Process}
\label{proce}

The merging process consists in inverting the
transformations, and then putting together the
resulting knowledge bases. Since we only have
the knowledge bases $K_i$'s {\em after} the
changes, we do not know for sure which
transformations are the ones to invert. Extending
the principles used for revision and arbitration,
we make some hypotheses about the kind of mistakes
that have been made. Considering only the most
likely possibilities, we are still left with a
number of possible scenarios. For each of them,
however, we know how to invert the transformations
and obtain the original knowledge bases $S_i$,
which can be then conjoined to get the maximum
possible information. What result is the merged
knowledge base in one of the possible scenarios
we assumed. Therefore, we have one knowledge base
for each scenario: since these are alternative
possibilities, the right way of combining them
is by disjunction.

Formally, we begin with the knowledge bases
$K_1, K_2, \ldots, K_n$, and make an assumption
about the transformations that have been used
to obtain them. Inverting these transformations,
we obtain $K_1', K_2', \ldots, K_n'$. If the
assumption about the transformations is correct,
the best way of merging them is simply by putting
them together, thus obtaining
$K=K_1' \wedge K_2' \wedge \cdots \wedge K_n'$.

On the other hand, this is only a possible scenario.
In another scenario, we may get a different result
of merging $K^1$, in another one we may have yet
another result $K^2$, etc. Since these are the results
of considering different alternatives we consider
equally likely, the final result of merging should
be the disjunction (logical or) of them.

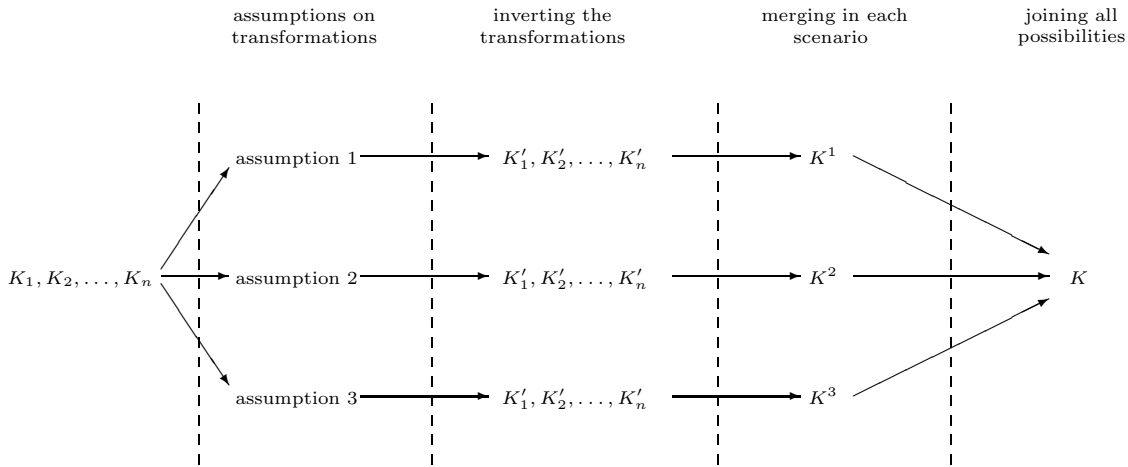
\begin{figure}[ht]
\begin{center}
\setlength{\unitlength}{2072sp}%
\begingroup\makeatletter\ifx\SetFigFont\undefined%
\gdef\SetFigFont#1#2#3#4#5{%
  \reset@font\fontsize{#1}{#2pt}%
  \fontfamily{#3}\fontseries{#4}\fontshape{#5}%
  \selectfont}%
\fi\endgroup%
\begin{picture}(13557,5478)(534,-4843)
\thinlines
\multiput(3151,-511)(0.00000,-233.51351){19}{\line( 0,-1){116.757}}
\multiput(5941,-511)(0.00000,-233.51351){19}{\line( 0,-1){116.757}}
\put(2701,-2581){\vector( 1, 0){810}}
\put(2701,-2671){\vector( 2,-3){810}}
\put(2701,-2491){\vector( 2, 3){810}}
\put(5086,-1141){\vector( 1, 0){1530}}
\put(5086,-2581){\vector( 1, 0){1530}}
\put(5086,-4021){\vector( 1, 0){1530}}
\multiput(9361,-511)(0.00000,-233.51351){19}{\line( 0,-1){116.757}}
\put(8821,-1141){\vector( 1, 0){1530}}
\put(8821,-2581){\vector( 1, 0){1530}}
\put(8821,-4021){\vector( 1, 0){1530}}
\multiput(12151,-511)(0.00000,-233.51351){19}{\line( 0,-1){116.757}}
\put(10981,-1141){\vector( 2,-1){2340}}
\put(10981,-2581){\vector( 1, 0){2340}}
\put(10981,-4021){\vector( 2, 1){2340}}
\put(2611,-2671){\makebox(0,0)[rb]{\smash{\SetFigFont{7}{8.4}{\familydefault}{\mddefault}{\updefault}$K_1, K_2, \ldots, K_n$}}}
\put(4321,-1231){\makebox(0,0)[b]{\smash{\SetFigFont{7}{8.4}{\familydefault}{\mddefault}{\updefault}assumption 1}}}
\put(4321,-2671){\makebox(0,0)[b]{\smash{\SetFigFont{7}{8.4}{\familydefault}{\mddefault}{\updefault}assumption 2}}}
\put(4321,-4111){\makebox(0,0)[b]{\smash{\SetFigFont{7}{8.4}{\familydefault}{\mddefault}{\updefault}assumption 3}}}
\put(4411,479){\makebox(0,0)[b]{\smash{\SetFigFont{7}{8.4}{\familydefault}{\mddefault}{\updefault}assumptions on}}}
\put(4411,209){\makebox(0,0)[b]{\smash{\SetFigFont{7}{8.4}{\familydefault}{\mddefault}{\updefault}transformations}}}
\put(7381,209){\makebox(0,0)[b]{\smash{\SetFigFont{7}{8.4}{\familydefault}{\mddefault}{\updefault}transformations}}}
\put(7381,479){\makebox(0,0)[b]{\smash{\SetFigFont{7}{8.4}{\familydefault}{\mddefault}{\updefault}inverting the}}}
\put(10621,-1231){\makebox(0,0)[b]{\smash{\SetFigFont{7}{8.4}{\familydefault}{\mddefault}{\updefault}$K^1$}}}
\put(10621,-2671){\makebox(0,0)[b]{\smash{\SetFigFont{7}{8.4}{\familydefault}{\mddefault}{\updefault}$K^2$}}}
\put(10621,-4111){\makebox(0,0)[b]{\smash{\SetFigFont{7}{8.4}{\familydefault}{\mddefault}{\updefault}$K^3$}}}
\put(10756,479){\makebox(0,0)[b]{\smash{\SetFigFont{7}{8.4}{\familydefault}{\mddefault}{\updefault}merging in each}}}
\put(10711,209){\makebox(0,0)[b]{\smash{\SetFigFont{7}{8.4}{\familydefault}{\mddefault}{\updefault}scenario}}}
\put(13591,479){\makebox(0,0)[b]{\smash{\SetFigFont{7}{8.4}{\familydefault}{\mddefault}{\updefault}joining all}}}
\put(13591,224){\makebox(0,0)[b]{\smash{\SetFigFont{7}{8.4}{\familydefault}{\mddefault}{\updefault}possibilities}}}
\put(13681,-2671){\makebox(0,0)[b]{\smash{\SetFigFont{7}{8.4}{\familydefault}{\mddefault}{\updefault}$K$}}}
\put(7651,-1231){\makebox(0,0)[b]{\smash{\SetFigFont{7}{8.4}{\familydefault}{\mddefault}{\updefault}$K_1', K_2', \ldots, K_n'$}}}
\put(7651,-2671){\makebox(0,0)[b]{\smash{\SetFigFont{7}{8.4}{\familydefault}{\mddefault}{\updefault}$K_1', K_2', \ldots, K_n'$}}}
\put(7651,-4111){\makebox(0,0)[b]{\smash{\SetFigFont{7}{8.4}{\familydefault}{\mddefault}{\updefault}$K_1', K_2', \ldots, K_n'$}}}
\end{picture}
\caption{The merging process}
\label{merge-process}
\end{center}
\end{figure}

Figure~\ref{merge-process} shows this process. Finding
and inverting the transformations are central steps of this
process: on the one hand, we should select as few possible
scenarios as possible to avoid a too weak result; on the
other hand, including too few possibilities may lead us
to neglect the one that really represents the state of
the world.

For simplicity, we replace these first two steps of the
process by the one of finding one (or more) $n$-tuples of
inverse transformations, one for each knowledge base.
Indeed, finding the transformations that have been
applied and inverting them can be formalized by the single
step of finding the transformations that lead from the
knowledge bases we have to the original ones; we call
them ``inverse transformations'' simply because they invert
the transformations that have been previously applied, but
they are still the transformations previously considered,
like variable substitution, etc.

In order to select one (or more) $n$-tuple of inverse
transformations, we define an ordering over all possible
$n$-tuples of sets of transformations. This way, we can
compare a possible scenario with another one, and tell
which one is the most likely.
A different and simpler model is that in which there is
one ordering for each knowledge base. We do not adopt this
model because mistakes in one knowledge base should be ranked
not only according to that source, but also as a result of comparing
it with the other knowledge bases. This is why we consider
an ordering ranking $n$-tuples rather than comparing
transformations locally, \ie, source by source.

This ordering may originate in different ways: it can be
part of the knowledge of each source (that is, each agent
has its own idea of the mistakes it likely makes), or it
can be an information the merger has (possibly based on
the meaning of the literals and other related knowledge),
or it is derived from the knowledge bases $K_i$'s using
some heuristics. In the first two cases, we can simply
assume that the ordering is given; the problem of obtaining
it from the knowledge bases is discussed
in the next session. Either way, in the rest of this section
we assume that this ordering is given. In particular, we
assume that $\R$ is a function that associates an integer
to each $n$-tuple of sets of transformations, giving
the likeliness they correct the mistakes in the
knowledge bases $K_1,\ldots,K_n$. As is common in belief
revision, we interpret a lower rank as an higher degrees of
likeliness, and therefore prefer $n$-tuples with the lowest
rank.

The set of possible transformation that may have been used
for generating $K_i$ from $S_i$ is defined as follows:
\eatpar

\[
\T(K_i) =
\{ \tau^h_{x,y} ~|~ y \in \var(K_i) ,~ x \not\in \var(K_i) \}
\cup
\{ \tau^g_x ~|~ x \not\in \var(K_i) \}
\cup
\{ \tau^p_x ~|~ \neg x \models K_i \}
\]

For example, $K_i$ may result from replacing $x$ with $y$,
and this is why the renaming of $x$ with $y$ is in the
the first part of $\T(K_i)$ only if $y$ is mentioned in $K_i$
while $x$ is not. The other parts of $\T(K_i)$ are motivated
in a similar way. The set $\T(K_i)$
is potentially infinite, as there are potentially
infinite possible variables $x \not\in \var(K_i)$.
For example, if $S_i=x \vee y$, and the source renamed
$x$ with $z$, it ends up with $K_i=z \vee y$. Inverting this
transformation amounts to deciding which name $z$ originally
had, and this is impossible by looking at $K_i$ only.
When a variable disappears from a knowledge base, like in
this case, we either use a variable that appears in another
knowledge base, or introduce a new one. This limits the set
of possible transformations: when we write $x \not\in \var(K_i)$
we assume that either $x$ is a variable occurring in some
other knowledge base, or $x$ is a new variable
created on purpose.

In order to invert the transformations, we define an inverse
relation $Inverse_i(\tau_1,\tau_2)$, which relates two
transformations $\tau_1$ and $\tau_2$ in such a way $\tau_2$
undoes the changes made by $\tau_1$ on the knowledge base
$K_i$. Note that $Inverse_i$ is indexed by $i$, thus making
this relation dependent on the considered knowledge base.
However, only the names of the variables in $K_i$ are really
needed. Also note that $Inverse_i$ is not a function, as
renaming and generalization cannot be uniquely
inverted. This relation is formally defined as follows.
\eatpar

\begin{eqnarray*}
Inverse_i &=&
\{ (\tau^h_{x,y}, \tau^h_{y,z}) ~|~
\tau^h_{x,y} \in \T(K_i) ,~ x \in \var(K_i) ,~ z \not\in \var(K_i) \}
\cup
\\
&&
\{ (\tau^g_x, \tau^p_y) ~|~ \tau^g_x \in \T(K_i) ,~ y \not\in \var(K_i) \}
\cup
\\
&&
\{ (\tau^p_x, \tau^g_y) ~|~  \tau^g_x \in \T(K_i) ,~ y \not\in \var(K_i) \}
\end{eqnarray*}

The relation $Inverse_i$ defines the set of {\em all} possible
inverse transformations on the knowledge base $K_i$. Since there
are too many such transformations, we also consider the ordering
that tells their degree of likeliness. This ordering is formalized
as a functions from $n$-tuples of sets of transformations to integers.
Formally, an integer is associated to each subset of
$Inverse_1 \times \cdots \times Inverse_n$. The idea is that each
subset of this set contains a set of transformations for each
knowledge base; implicitly, it tells the mistakes that have been
done. The ordering simply tells the degree of likeliness of
these mistakes. We denote this function as $\R$.

This ranking makes the process of merging possible.
As it is common in belief revision, we consider all possible
changes to the knowledge bases, select only the ones that lead
to the expected result ($A$ should be derivable but $\neg B$
should not), and then use the ranking to further reduce the
set of possibilities.

In order to define the first step (selection of transformations),
we have to specify, for each $n$-tuple of sets of
transformations, what is the resulting knowledge base.
Let therefore $\l \L_1, \ldots, \L_n \r$ be this n-tuple, where
$\L_i \subseteq
\{ \tau ~|~ \exists \tau' ~.~ \l \tau',\tau \r \in Inverse_i \}$.
The result of applying the transformations in $\L_i$ to
$K_i$ is as follows:\eatpar

\[
\I_{\L_i} (K_i) = \tau_1(\ldots(\tau_n(K_i))
\mbox{~~~ where ~~~} 
\L_i = \{ \tau_1, \ldots, \tau_n \}
\]

We extend this operator to tuples of set of transformations
and to tuples of knowledge bases as follows.\eatpar

\[
\I_{\l \L_1, \ldots, \L_n \r} \K = 
\bigwedge
\I_{\L_i} (K_i)
\]

This is the result of merging only if $\L_1, \ldots, \L_n$
is known to be the way in which the transformations have to
be inverted, or it is the only way in which both the
constraint on $A$ and the constraint on $B$ can be
satisfied. Usually, this is not the case, so we have to use
the ranking $\R$ to make a selection.

The transformations we consider are the minimal ones among
those making the result of merging to imply $A$ but not to
imply $\neg B$. Minimality is defined using the
ranking.\eatpar

\begin{eqnarray*}
\mbox{$\M_\tau$}
&=&
\min_\R(\{ \l \L_1,\ldots,\L_n \r ~|~ 
\bigwedge_{i=1,\ldots,n}
     \I_{\l \L_1, \ldots, \L_n \r} \K \models A
\mbox{~~~ and ~~~}
\bigwedge_{i=1,\ldots,n}
     (\I_{\l \L_1, \ldots, \L_n \r} \K) \wedge B \not\models \bot \},
)
\end{eqnarray*}

This formula defines a set of transformations for each
source. Clearly, there is no warranty that such a minimum is
unique. The merger applies each set of possible
transformations, and disjoins the results: \eatpar

\[
\I^A_B(\K) = 
\bigvee_{\L \in \M_\tau} \I_{\L} \K
\]

By construction, $\I^A_B(\K)$ implies $A$ simply because it
is a disjunction of terms, each implying $A$. For the same
reason, since each term is consistent with $B$, so is the
result of merging.

\section{Selection Heuristics}
\label{heuri}

The merging process outlined in the last section depends on
which the most likely transformations are. So far, we simply
assumed the knowledge of the ordering $\R_i$, either because
it is an additional information the agents have, or because
it is known to the centralized merger. However, the case in
which no additional information, besides the knowledge
bases, is known is also important. In this section we
consider the case in which no information about the
likeliness of mistakes is given, and the ordering must be
drawn from the knowledge bases $K_i$.

While it is always theoretically possible to select all
possible transformations that satisfy the constraints $A$
and $B$, these transformations may be too many to give
useful information. Indeed, the more the possible considered
scenarios are, the weaker the resulting knowledge base is,
and the number of possible scenarios may be very large even
for very simple knowledge bases. For example, the knowledge
base $K_1=x \rightarrow y$ may result from the renaming of
$z$ to $x$, or from the particularization of $y$ (i.e.\  we
incorrectly assumed that $y$ holds only when $x$ is true),
or from the generalization of $x \wedge z \rightarrow y$,
etc. A first selection criteria is that we only accepts sets
of transformations that produce a knowledge base that
satisfies the upper and the lower bounds. This, however, may
be still too weak a constraint to limit the number of
transformations.

For this reason, we also assume some minimality criteria;
namely, we assume that as few mistakes as possible have been
made while producing $K_i$ from $S_i$. In a sense, this is
the minimal change principle in disguise: assuming a minimal
number of mistakes, we still consider a minimal number of
(inverse) transformations to be applied to the knowledge
bases. On the other hand, the minimal change principle in
this form is not a first principle any longer, but only a
consequence of a more general assumption.

The principle of minimizing the number of
mistakes/transformations, however, may still be not enough,
that is, the number of possible scenarios may still be too
high. Therefore, we use the knowledge bases to further limit
the number of possible alternatives. In this section, we
present a selection heuristics that is based only on the
knowledge bases. We assume that no further information is
given about the meaning of literals, the likeliness of
mistakes, etc. and that we cannot perform any
information-gathering actions (a common assumption in belief
revision, less in the real world.)

Another problem of the merging process is that some
transformations cannot be inverted uniquely. In
particular, knowing that $K_i = \tau^g_x(S_i)$ does not allow
to derive $S_i$. In such cases, we simply assume that
$S_i=\tau^p_x(K_i)$. This is equivalent to assuming that
$\tau^g_x$ is only applied to formulae like
$x \rightarrow F$, i.e., having $x$ as a precondition.

In order to define this ranking $\R$, we observe that it
only needs to rank the transformations according to their
plausibility, regardless of whether they lead to satisfy the
lower and upper bounds of merging: it is the merging process
that enforces these constraints to be satisfied.

The ranking $\R$ is based on (besides assuming a minimal
number of mistakes,) assuming that the initial knowledge
bases $S_i$'s are similar to each other. Therefore, the best
inverse transformations are those making the resulting
knowledge bases $K_i'$ as similar to each other as possible.

Examples justifying this way of operating are easy to find:
if a knowledge base is identical to another one except for a
different variable name, the change of the name is
intuitively the most reasonable action to do before
integrating the two knowledge bases.

This example can be generalized to the case in which
applying a transformation to $K_i$ makes it equal to $K_j$:
this transformation is likely to be the inverse of the one
that changed $S_i$ to $K_i$. In this case, $S_i=S_j$, but is
not always the case. To make this criteria to have general
applicability, we need a way for applying it even when the
two knowledge bases cannot be made identical. To this aim,
we measure the similarity between knowledge bases, and trade
off between the number of inverse transformations and the
degree of similarity of the resulting knowledge bases. We
therefore need a way for measuring the similarity between
two knowledge bases, and then a way for combining this
measure with the number of changes needed to make the
knowledge bases similar.

The measure of similarity can be defined either syntactically
or semantically; we define a semantical measure. There are two reasons
for this choice: first, it is possible to express the same knowledge
in different ways (so that $S_i$ and $S_j$, while identical in
their sets of models, are syntactically different); second,
each source may have further changed the syntactic form of its
knowledge base to suit its purposes.

Let $K_1$ and $K_2$ be two knowledge bases, and let $\mod(K_1)$
and $\mod(K_2)$ be their sets of models. The measure of similarity
should grow as the size of the intersection $\mod(K_1) \cap \mod(K_2)$,
and as the intersection of their complements
$\mod(\neg K_1) \cap \mod(\neg K_2)$. The total size of
these two sets is in fact equal to $|\mod(K_1 \equiv K_2)|$.
The degree of similarity should also decrease with the number of
models that satisfy only one formula, that is, the size of
$\mod(K_1 \not\equiv K_2)$. A possible choice is the linear
combination of these two measures:
\eatpar

\[
\delta(K_1,K_2) = 
| \mod(K_1 \equiv K_2) | - | \mod(K_1 \not\equiv K_2) | =
2* | \mod(K_1 \equiv K_2) | - | \mod(\true) | 
\]

This function is in practice the same as
$| \mod(K_1 \equiv K_2) |$. Another possibility
is that of using a quotient: $
\delta(K_1,K_2) =
| \mod(K_1 \equiv K_2) | / | \mod(K_1 \not\equiv K_2) |$.

Having defined the measure of similarity $\delta$ of two
knowledge bases, we can now combine it with the number of
transformations to define the ranking. Let us therefore
consider a specific $n$-tuple $\l \L_1,\ldots,\L_n \r$. The
knowledge bases generated by the transformations are
$\I_{\L_i}(K_i)$. We compare them using $\delta$ and the
number of transformations in each set $\L_i$. We use a
simple linear combination of these two measures.

\[
\R(\l \L_1,\ldots,\L_n \r)=
\sum_{K_i, K_j} \log(\delta(\I_{\L_i}(K_i), \I_{\L_j}(K_j))+1)
+
\sum |\L_i|
\]

The logarithm is used to make the measure of similarity and
the number of transformations to be on the same scale:
without it, the measure of similarity can be exponentially
large, thus making the contribution of the number of
transformations irrelevant. We used a logarithm (instead of
a multiplying factor) because a difference of distances
should be less important when the number of different models
is high: the difference between one model and two is more
important than the difference between 1000 and 1001.

This ranking $\R$ defines a measure of goodness of
transformations, and therefore completes the merging
process outlined in the previous section: given the
knowledge bases $K_i$'s, we can now tell exactly what
the result of merging is.

A problem of this ranking, however, is that it is based on
assuming that all knowledge bases $K_i$ derives from similar
knowledge bases $S_i$ by applying some, equally likely,
transformations. While the equal likeliness is the natural
result of assuming no information about the likeliness of
transformations, the assumption that the $S_i$'s are similar
is questionable. In particular, it may be more reasonable to
assume that each knowledge base is ``targeted'' to a
different subject. Indeed, it is likely that each source
uses the knowledge base for a specific purpose; as a result,
the knowledge bases may contain only information about some
specific subjects.

To take this consideration into account, we do not measure
how similar the knowledge bases are, but how similar they
are when restricted to a subset of variables. Namely,
let $K^Y$ be the restriction of the formula $K$ to the
variables in $Y$. The difference between two formulae $K_1$
and $K_2$ is:
\eatpar

\[
\delta(K_1,K_2) =
\sum_{\emptyset \not= Y \subseteq X}
\frac{\delta(K_1^Y,K_2^Y)}
{|X|-|Y|+1}
\]

This is how we formalize the assumption that the result of
the inverse transformations may be a formula that is similar
to the other one only for a subset of its variables. The
quotient is defined in such a way to avoid a difference in
the case $|Y|=1$ to count the same as in the case $Y=X$.

\comment

Properties of this operator:

1. works in the correct way if all initial knowledge
  bases are equal but differ for a single transformation
  applied to it

2. does not introduce consequences there should not be
  there? (skeptical operator)

3. other?

\endcomment

\comment
if a knowledge base is the same as ten knowledge bases
to which a transformation has been applied, then the
ranking of the other ones should increase (it is more
likely that a single transformation has been applied).
\endcomment

\section{The Renaming Merging Operator}
\label{compl}

The ranking defined in the previous section allows for
determining the result of merging from the knowledge bases
alone, without any additional information. In the belief
revision terminology, this is a {\em merging operator}, as
opposed to {\em merging schemas}, which require some
additional information such as ranking, preferences, etc.
(they are called schemas because they are the backbones of a
merging process, but something has to be added to make them
complete merging operators.)

The operator defined in the last section allows for checking
the validity of properties that should hold for the merging
process. However, the number of possible transformations
make the operator quite complicated. We therefore make the
simplifying assumption that the only mistakes are those
involving renamings. The set of possible transformations is
therefore defined as follows.

\begin{definition}

Given a set of variables $X$, a permitted inverse
transformation is a substitution $X/Y$ in which
each $x_i$ is either substituted with another
variable in $X$, or it is renamed as the new
variable $x_i'$.

\end{definition}

This definition forbids the proliferation of new variables:
if we have to replace $x_i$ with a new variable, we are
forced to name it $x_i'$. This rule limits the number of
choices while renaming variables. Intuitively, if we have to
change the name of a variable, this rule allows not to care
about the name of the new variable.

The merging operator is based on a particularization of the
general model of the sources; namely, the only considered
transformations are renamings. Therefore, in order to define
a specific merging operator, we only need an ordering over
the renamings. Assuming all transformations equally likely
we get a merging operator we call {\em Renaming Merging with
Equal Likeliness Operator}, or RMEL for short. For the sake
of clarity, we only consider two knowledge bases, as is
common in the merging/arbitration literature.

\begin{definition}

The {\em Renaming Merging with Equal Likeliness Operator}
$*_{RMEL}^{A,B}$ associates any two knowledge bases $K_1$
and $K_2$ to another knowledge base $K_1 *_{RMEL}^{A,B} K_2$
defined as follows:

\[
K_1 *_{RMEL}^{A,B} K_2
=
\bigvee_{\l Y, Z \r \in PIT}
K_1[X/Y] \wedge K_2[X/Z]
\]

\noindent
where $\l Y,Z \r \in PIT$ if and only if $X/Y$ and
$X/Z$ are permitted inverse transformations that
satisfy $K_1[X/Y] \wedge K_2[X/Z] \models A$ and
$K_1[X/Y] \wedge K_2[X/Z] \wedge B \not\models \bot$
and are of minimal combined size (that is, the
size of $Y$ plus that of $Z$ is minimal.) $X$ is a
subset of the variables in $K_1$, $K_2$, $A$, and $B$.

\end{definition}

In this definition, we consider renamings to the variables
in $K_1$ and in $K_2$ that satisfy the bounds $A$ and $B$.
Using only permitted inverse transformations reduces the
number of disjuncts in the definition. Indeed, for each
variable in each of the two knowledge bases, we can either
substitute it with another variable in $X$, or with a new
variable not appearing anywhere else. The use of new
variables is necessary as the two knowledge bases may use
the same variables for different facts, so that either one
or both of them have to be renamed. Using only permitted
inverse transformations we avoid the problem of having to
consider transformations that differ only for the name of
the new variables, since the name of new variables is
defined uniquely. On the other hand, permitted
transformations are liberal enough to allow for making the
alphabets of $K_1[X/Y]$ and $K_2[X/Z]$ disjoint (just
substitute each $x_i$ with $x_i'$ in $K_1$, and make no
changes to $K_2$.) This may be necessary when the two
knowledge bases use exactly the same variables to represent
completely different facts.

The rule of minimality excludes transformations that
introduce renamings that are not justified. This particular
ordering is the one that reduces the number of renamings the
most, but other rules can be used instead: minimality
w.r.t.\  set containment, minimal size of $Y$ and $Z$
considered separately, user supplied ranking of
transformations, etc.

Let us consider some properties of this operator. We
assume that both $A$ and $B$ are consistent, and
do not contradict each other. This is important:
for example, if $A=a$ and $B=\neg a$,
there is no way to make $A$ implied and $B$
consistent at the same time. We therefore assume that
$A$, $B$, and $A \wedge B$ are all consistent. A good
property this merging operator should have is that of
success, that is, it should produce a meaningful result.
In our case, since merging is
defined as a disjunction of possible hypotheses, this
amounts to checking whether the resulting knowledge
base is consistent. Unfortunately, this may
not be the case, as the following example shows:

\begin{eqnarray*}
K_1 &=& \neg x_1 \\
K_2 &=& \neg x_2 \\
A &=& x_1 \\
B &=& \top
\end{eqnarray*}

The problem here is that the two knowledge bases both
tell that something is false, while we wanted a variable
to be true after the merging, as $A=x_1$. The problem
could be overcome by considering transformations involving
negative literals, but this is quite unintuitive in this case:
if we assume that the only problem is that we are giving
the wrong name to a fact, we cannot infer that a fact is
true from a statement saying that a fact is false.

The reason of why we cannot get success in this example is
that the operator is based on assuming that the knowledge
bases are obtained by renamings of correct ones, but the
knowledge bases $K_1$ and $K_2$ of this example contradict
this assumption. Indeed, $K_1 = \neg x_1$ cannot be the
result of changing a name to a knowledge base that implies
$x_1$. Obtaining a consistent result from the knowledge
bases above would therefore be counterintuitive, as the
merging operator would be saying that the assumption on the
transformations (only name changes are possible) is
consistent with the available data, while in fact it is not.

This example shows that we cannot expect the merging
operator to work correctly even when the assumptions it is
based on do not hold. On the contrary, the properties of
this operator have to be checked with respect to two
knowledge bases $K_1$ and $K_2$ that actually result from
renaming some variables in two knowledge bases $S_1$ and
$S_2$, both consistent with $B$ and both implying $A$.

If this is the case, the transformations can be inverted,
and therefore the bounds $A$ and $B$ can be satisfied. It
does not matter that the inverse transformation is not
unique: to achieve derivability of $A$ and consistency with
$B$, all that is needed is that there is at least a pair $\l
Y,Z \r$ such that $K_1[X/Y] \wedge K_2[X/Z] \models A$ and
$K_1[X/Y] \wedge K_2[X/Z] \wedge B \not\models B$. All other
disjuncts involved in the definition (if any) are consistent
with $B$, and therefore their disjunction is consistent with
$B$ as well. The upper bound $A$ is satisfied for the same
reason: since each element of the disjunction implies $A$,
all of its models are models of $A$.

A second property that we wish to obtain is
that the original knowledge is correctly, even
if not completely, recovered. This is to say that, if
$S_1 \wedge S_2 \not\models C$, then the result of
merging $K_1$ with $K_2$ should not imply $C$ either.
However, this is not always the case, as the following
example shows.

\begin{eqnarray*}
K_1 &=& \hbox to 1cm{$x_1$\hfill}  (S_1 =x_2) \\
K_2 &=& \hbox to 1cm{$\top$\hfill}  (S_2 = \top) \\
A &=& \top \\
B &=& \top
\end{eqnarray*}

This example clearly shows a problem that has been already
mentioned in the introduction: if we have no way to realize
that a mistake has been made, then there is no way to
recover from it. In this case, assuming that both knowledge
bases are free of mistakes is not inconsistent with the
bounds $A$ and $B$. Therefore, $K_1 \wedge K_2 = x_1$ is the
result of merging simply because we have no reason to assume
that a name change is necessary. This conclusion is
incorrect, as $x_1$ is not a consequence of $S_1 \wedge S_2
= x_2$. This example also shows the obvious fact that we
cannot enforce completeness either: $x_2$ is a consequence
of the original knowledge base, but is not a consequence of
the result of merging.

The fact that we cannot always recover the original
knowledge bases, however, it is not unique to this operator.
Even in the ``mistake of value'' assumption (that is, in
``traditional'' belief revision operators), the way the
result of merging is related to the real world is
conditioned to the validity of the minimal change principle.
To make a concrete example, if our real world is $a \wedge
b$, and we have to revise $K=\neg a$ to $P=b$, we will
always get the incorrect conclusion $\neg a$. This is simply
because:

\begin{enumerate}

\item there is no evidence we need to make any change;

\item we commit to the principle of making as few
changes as possible.

\end{enumerate}

In our scenario, we do not have any evidence that makes
us thinking that a mistake has been made, and we therefore
assume that the knowledge bases are correct. Making any
other choice without any additional justifying information
would be unmotivated.

We now consider the operator obtained by adding the
ranking over transformations defined in the previous
section. The definition of ranking specializes to
the case of renamings only as follows.

\begin{definition}

The degree of a permitted inverse transformation
$X/Y, X/Z$ w.r.t.\  $K_1$, $K_2$, $A$, and $B$
is given by the following formula:

\[
\R(\l X/Y, X/Z \r) =
|X/Y| + |X/Z| +
\log(\delta(K_1[X/Y], K_2[X/Z]))
\]

\end{definition}

In words, this ranking combines the number of name changes
with the similarity of the knowledge bases after the changes
(the similarity measure $\delta$ can be defined as shown in
the previous section.) In this case, we have used a simple
linear combination, but other combinations are possible (for
example, we can first consider the number of changes, and
then the similarity only in case of ties.)

\begin{definition}

The {\em Renaming Merging Operator}
$*_{RM}^{A,B}$ associates with any two knowledge bases $K_1$
and $K_2$ another knowledge base $K_1 *_{RM}^{A,B} K_2$
defined as follows:

\[
K_1 *_{RM}^{A,B} K_2
=
\bigvee_{\l Y, Z \r \in MPIT}
K_1[X/Y] \wedge K_2[X/Z]
\]

\noindent
where $\l Y,Z \r \in MPIT$ if and only if $X/Y$ and
$X/Z$ are minimal permitted inverse transformations
w.r.t.\ $K_1$, $K_2$, $A$, and $B$.

\end{definition}

This operator differs from the previous one only
in that the similarity between the two knowledge
bases is taken into account, and it is in the same
degree as the number of substitutions.

The same drawbacks of the operator with equal likeliness
appear here. The difference is that, using an ordering, we
select less transformations. Thus, we have less terms in the
disjunction, and therefore the result of merging can be
logically stronger.

\section{Complexity Results}

In this section, we consider the complexity of inference
for the renaming merging with equal likeliness operator.
Formally, given $K_1$, $K_2$, $A$, $B$, and $Q$, we want 
to check whether $Q$ is implied by the merge of $K_1$
with $K_2$, where $A$ and $B$ are the upper and lower
bound, respectively.

\comment

wrong proof of hardness of consistent
renaming implication, dropped

\endcomment

\begin{theorem}

The problem of checking whether
$K_1 *_{RMEL}^{A,B} K_2 \models Q$
is \P{2}-hard, and is in \Dlog{3}.

\end{theorem}

\proof Membership: finding the size of the minimal renamings
that make $K_1$ and $K_2$ consistent with $B$ and not with
$\neg A$ can be done with a logarithmic number of queries to
an oracle that checks the existence of a substitution that
satisfies both constraints (this oracle must be in the second
level of the polynomial hierarchy due to the upper bound: it
has to check the existence of a substitution such that the
resulting knowledge bases imply $A$.) 

Using the minimal size of substitutions, all is needed is to
check whether the knowledge bases imply $Q$ using all
substitutions of minimal size that satisfy both constraints.

\

Hardness is proved by reduction from $\forall\exists$QBF.
We prove that $\forall X \exists Y . F$ is valid if and
only if $K_1 *_{RMEL}^{A,B} K_2 \models Q$, where $Q=a$ and

\begin{eqnarray*}
K_1 &=& a \wedge x_1 \wedge \cdots \wedge x_n \\
K_2 &=& \neg a \wedge \neg x_1 \wedge \cdots \wedge \neg x_n \\
A &=& a \vee (\neg F[Y/Y_1] \wedge \cdots \wedge \neg F[Y/Y_{n+1}]) \\
B &=& \top
\end{eqnarray*}

In order to satisfy the lower bound, the substitutions
must make $K_1$ and $K_2$ consistent. This is only possible
by changing the name of each variable in $\{a,x_1,\ldots,x_n\}$
either in $K_1$ or in $K_2$. This way, putting together
$K_1$ and $K_2$, we obtain a formula that contains exactly
one literal between $x_i$ and $\neg x_i$ and one literal
between $a$ and $\neg a$, that is, a formula having exactly
one model over variables $X \cup \{a\}$.

Changing the names this way is necessary to satisfy the
lower bound $B$. We can also prove that $n+1$ name changes
are sufficient to satisfy the upper bound $A$. The
substitutions that rename $a$ in $K_2$ are such that $a$
is implied by $K_1$ and $K_2$ after the renaming; therefore,
$A$ is implied as well. As a result, exactly $n+1$ variable
name changes are needed to make both constraints satisfied.
In particular, each variable in $X \cup \{a\}$ has to be
renamed in either $K_1$ or $K_2$: if the knowledge base that
results satisfies $A$, then this substitution is considered.

Let us first consider the case in which all variables
in $K_1$ and $K_2$ are replaced with new ones. In order
to make $K_1$ and $K_2$ consistent, we have to rename
any variable in $\{a,x_1,\ldots,x_n\}$ either in $K_1$
or in $K_2$. After the change, $K_1$ and $K_2$ is a
knowledge base with exactly one model. The substitution
is considered only if $A$ is implied by this model. By
construction, $A$ is implied only if either $a$ is true,
or the value of the variables $\{x_1,\ldots,x_n\}$ satisfy
$\neg F$ for all possible values of $Y$. As a result,
a substitution that makes $a$ false satisfies the upper
bound if and only if the corresponding evaluation of the
variables $X$ falsifies $F$ for any possible assignment
of the variables $Y$. As a result, $Q=a$ is implied if
and only if such assignments do not exist, that is, for
all values of $X$, there is a value of $Y$ that satisfy
$F$.

The reduction is proved only if we restrict
to substitution changing the name of a variable with
a new name. Let us now consider the other substitutions.
If a variable $x_i$ is renamed to $x_j$, all is said
above still holds (as the variable $x_i$, in a way or
another ``disappears'' from the knowledge base, and
therefore it is set to the value it has in the other
one.) The only substitutions that cause problems are those
changing the value of $x_i$ (or $a$) into a variable in
$Y$. Let for example consider the case in which the
substitution $x_1/y_1$ is applied to $K_1$. Then,
$x_1$ is set to false in the resulting merging.
At the same time, however, $y_1$ is set to true as well.
This is a problem if $\neg F$ is not satisfied by
the values of $X$ alone, but it is if $y_1$ is true: if $a$
is set to false, we obtain that $Q=a$ is not implied any
more, while we know that the partial evaluation of $X$
does not satisfy $F$. This is why $A$ contains $n+1$
copies of $F$: however we change the names of variables
in $X$ to variables in $Y$, the upper bound $A$ always
contain a copy of $F$ whose variables in $Y_i$ are not
mentioned in $K_1$ and $K_2$ after renaming. This ensures
that $A$ can only be derived if the partial evaluation
of $X$ falsifies $F$.~\qed

\section{Conclusions}
\label{concl}

The contribution of this paper is in the approach taken,
rather than the proposed specific belief revision method.
Starting from a very general model of the integration
domain, we have shown that the existing semantics for
knowledge integration correspond to a specific assumption.
In this model, the sources get the knowledge bases they have
by a process of acquisition that is prone to errors;
previous integration semantics correspond to the assumption
that mistakes are of a specific kind (which we called
``mistakes of value''). Other mistakes are considered in
this paper, leading to completely new integration semantics.
The work reported here is still preliminary, as the
properties of merging in the new models have not yet been
fully investigated (comments and suggestions are welcome.)

New issues come from further generalizing this model. For
example, we have only considered the case in which all
knowledge bases are propositional. For first order logic,
new interesting cases arise: a form of generalization is to
transform $P(a)$ into $\forall x. P(x)$; the opposite of
particularization is also interesting; subject
misunderstanding is in this context different (it is the
change of a constant, not the change of a literal), etc.

Other issues arise from comparing the approach taken here
with ``classical'' belief revision. In the usual
formalization of belief revision, the knowledge expressed by
each source is actually a set of preferences, rather than
simply a knowledge base. This is because each agent involved
in the merging process not only has some beliefs, but also
acknowledges the possibility that they may be indeed false.
As a result, it also has a measure of preference (degree of
belief) over all facts it considers to be false, generating
an ordering over the possible worlds.

Modeling merging with the assumption of mistakes, such
ordering cannot be used. As it is clear from the heuristics
presented, it is impossible to express merging as a merging
of ranking, as the most likely transformations of each
source depend on the other knowledge bases. It is also true
that we could consider a more sophisticated model accounting
both rankings (expressing the measure of likeliness of
worlds according to each agent) and mistakes (that each
agent did while getting its knowledge).

Finally, let us briefly discuss the computation issues. The
result of Section~\ref{compl} shows that the proposed
semantics is computationally harder than the propositional
calculus, as expected. Nevertheless, it is not much harder
than most of the revision operators, that are \P{2}\
complete \cite{eite-gott-91-b}. A simplified definition has
been used, but it seems unlikely it did reduce complexity
much.


\bibliographystyle{plain}

\end{document}